\documentclass[runningheads]{llncs}
\usepackage{graphicx}
\usepackage{url}
\usepackage{multirow}
\usepackage{subcaption}

\begin{document}
\title{Web Table Classification\\based on Visual Features}
%
%
\author{Babette B\"uhler\inst{1} \and
Heiko Paulheim\inst{2}\orcidID{0000-0003-4386-8195}}
\authorrunning{B. B\"uhler  and H. Paulheim}
%
\institute{Hector Research Institute of Education Sciences and Psychology,\\University of T\"ubingen, T\"ubingen, Germany\\
\email{babette.buehler@uni-tuebingen.de}\\
\and
Data and Web Science Group, University of Mannheim, Mannheim, Germany\\
\email{heiko@informatik.uni-mannheim.de}
}
\maketitle              

\begin{abstract}
Tables on the web constitute a valuable data source for many applications, like factual search and knowledge base augmentation. However, as genuine tables containing relational knowledge only account for a small proportion of tables on the web, reliable genuine web table classification is a crucial first step of table extraction. Previous works usually rely on explicit feature construction from the HTML code. In contrast, we propose an approach for web table classification by exploiting the \emph{full} visual appearance of a table, which works purely by applying a convolutional neural network on the rendered image of the web table.
Since these visual features can be extracted automatically, our approach circumvents the need for explicit feature construction.
A new hand labeled gold standard dataset containing HTML source code and images for 13,112 tables was generated for this task. Transfer learning techniques are applied to well known VGG16 and ResNet50 architectures. The evaluation of CNN image classification with fine tuned ResNet50 (F1 93.29\%) shows that this approach achieves results comparable to previous solutions using explicitly defined HTML code based features. By combining visual and explicit features, an F-measure of 93.70\% can be achieved by Random Forest classification, which beats current state of the art methods.


\end{abstract}

\section{Introduction}
The world wide web constitutes the worlds largest freely available source of information covering almost every topic area. Especially web tables are of interest in this respect, because they present knowledge in a structured and concise form. They have been successfully employed as a data source in areas such as factual search \cite{yin_facto_2011}, entity augmentation \cite{yakout_infogather_2012} and knowledge base augmentation \cite{lehmberg_mannheim_2015}.

Before web tables can be employed as a powerful knowledge resource they have to be extracted. As web pages are built using the Hyper Text Markup Language (HTML), the intuitive approach to locate a table is via the \texttt{$<$table$>$} tag. Crestan and Pantel \cite{crestan_web-scale_2011} suggest that, based on their investigation of table type distribution on the web, an overwhelmingly large proportion of  88\% of tables defined by the \texttt{$<$table$>$} tag are layout tables, used for navigation or formatting purposes. Consequently, the very first step in table processing is the identification of genuine web tables presenting relational knowledge.

\begin{figure}[!tbp]
	\begin{center}
	    \begin{subfigure}[b]{.45\textwidth}
			\includegraphics[width=\textwidth]{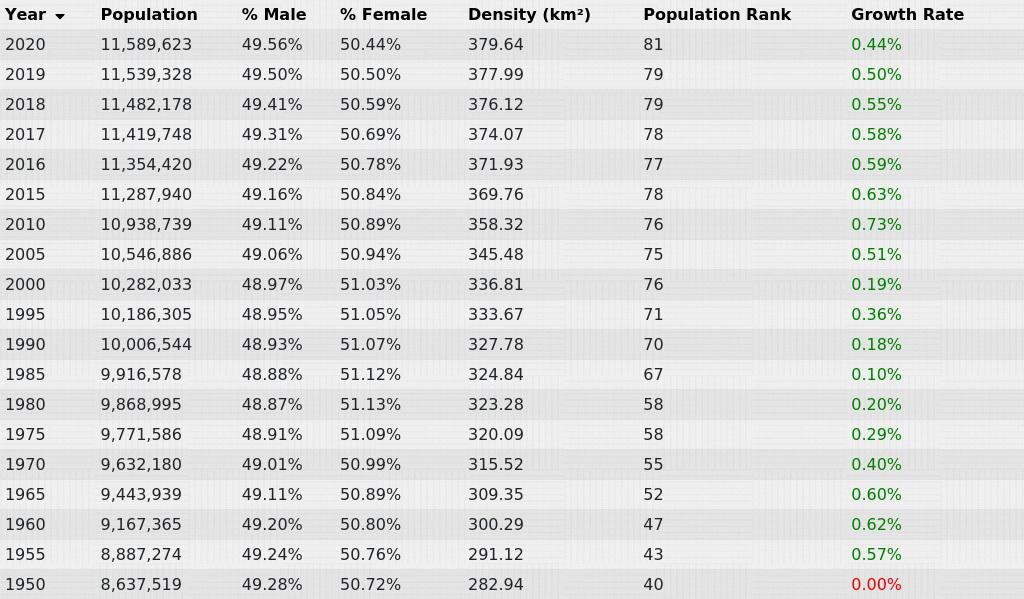}
			\caption{Example picture genuine table vertical listing}
			\label{fig:example_genuine1}
		\end{subfigure}
		\begin{subfigure}[b]{.45\textwidth}
				\includegraphics[width=\textwidth]{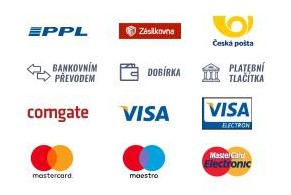}
				\caption{Example picture layout table formatting}
				\label{fig:example_layout1}
		\end{subfigure}

		\begin{subfigure}[b]{.45\textwidth}
			\includegraphics[width=\textwidth]{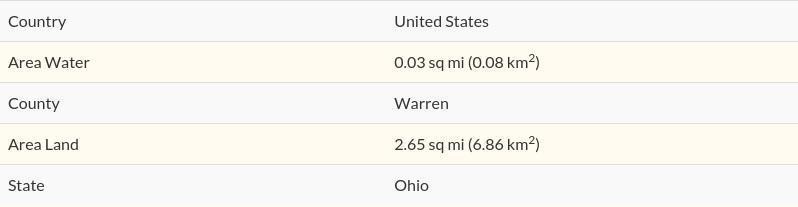}
			\caption{Example picture genuine table attribute/value}
			\label{fig:example_genuine2}
		\end{subfigure}
		\begin{subfigure}[b]{.45\textwidth}
			\includegraphics[width=\textwidth]{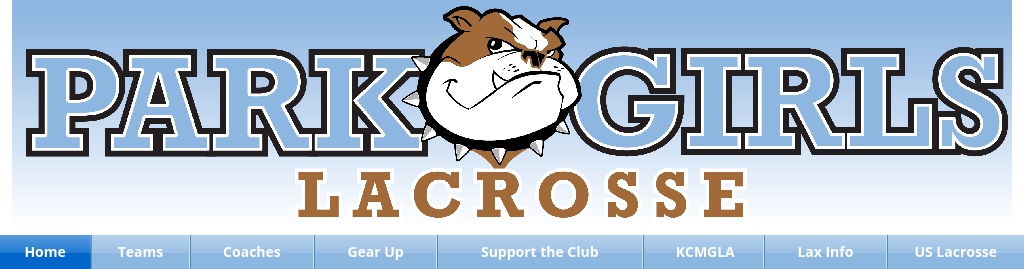}
			\caption{Example picture layout table navigating}
			\label{fig:example_layout2}
		\end{subfigure}
	\end{center}
	\caption{Examples for layout and genuine tables}
	\label{fig:examples}
\end{figure}

Figures~\ref{fig:example_layout1} and~\ref{fig:example_layout2} show the usage of HTML table elements for layout purposes, i.e., for arranging logos in a grid, and for a navigation bar. In contrast, figures~\ref{fig:example_genuine1} and~\ref{fig:example_genuine2} show genuine tables, once as a relational table and once as a table of key/value pairs referring to the same entity.

Previous research has worked on identifying genuine tables by applying extensive heuristic filter rules \cite{chen_mining_2000,penn_flexible_2001} and by using machine learning with explicitly defined features that describe the structure and genuine of a respective HTML table \cite{cafarella_uncovering_2008,eberius_building_2015,wang_machine_2002}. 
Most approaches extract those features from the HTML code. Exceptions are Cohen et al. \cite{cohen_flexible_2002} and Gatterbauer et al. \cite{gatterbauer_towards_2007} who employed quasi-rendered representations. However, the HTML code encodes the visual appearance of tables in the browser in a complex and indirect fashion. In contrast to this, it is in most cases intuitive for human viewers to distinguish layout from genuine tables in the rendered display of a website by their visual characteristics.

This paper proposes to automatically extract visual features by using convolutional neural networks from the rendered image representation of web tables. Two different strategies to the use of Convolutional Neural Networks (CNNs) for web table classification are presented. First, pre-trained and fine tuned CNNs are directly used to classify web tables, employing a dense classification layer. Further, pre-trained CNNs are used as standalone feature extractors and the extracted visual features, as well as their combination with explicitly defined features by Eberius et al. \cite{eberius_building_2015} are employed to train a Random Forest classifier. For each of the proposed approaches transfer learning and fine-tuning techniques for two well known deep learning architectures, VGG16 and ResNet50, with weights pre-trained on the ImageNet data, are tested.

The rest of this paper is structured as follows. Section~\ref{sec:related} outlines related work. Section~\ref{sec:approach} introduces our approach, which is analyzed in a set of experiments in section~\ref{sec:experiments}. We conclude with a summary and an outlook on future work.
 
\section{Related Work}
\label{sec:related}

A range of research has been conducted in the area of discriminating genuine and layout tables on the web. The approaches can be roughly divided into three strands, i.e., heuristic rules, machine learning using HTML based features, and visual representation based features.

\subsection{Heuristic Filter Rules}
Chen et al. \cite{chen_mining_2000} propose a set of filter rules and the use of cell similarity measures to detect genuine tables. They filter tables containing less than two cells or containing too many hyperlinks, forms, and figures. For the remaining tables, the similarity of neighboring cells is considered to tell genuine and layout tables apart. On a test dataset of 3,218 tables, they report an F-measure of 86.5\%.

In a similar approach, Penn et al. \cite{penn_flexible_2001} propose a set of heuristics to distinguish genuine and layout tables. Following these heuristics, genuine tables do not contain other tables, lists, frames, forms, images, have multiple rows and columms, more than one non-text-level-formatting tags, or less than a minimum amount of words. On a dataset extracted from 75 news websites, the authors report 86.3\% precision and 89.8\% recall. 

\subsection{Machine Learning Using HTML Based Features}
Wang and Hu \cite{wang_machine_2002} apply machine learning techniques for the genuine versus non-genuine table classification tasks. They propose a large set of layout and genuine type features, as well as a word group feature to train Decision Tree classifiers and Support Vector Machines (SVM). Layout features, e.g., the average number of rows or columns, were adopted to depict structural information, while genuine type features describe the type of the cells. On a test dataset consisting of 11,477 leaf tables they achieve a F-measure of 95.89\% for SVM with a RBF kernel. Additionally, it was compared to the rule based system proposed by Penn et al. \cite{penn_flexible_2001} which achieved a F-measure of 61.93\% and a F-measure of 87.63\% after ruling out the cell length threshold of four, on the same data.

Carafella et al. \cite{cafarella_uncovering_2008} employ a combination of heuristic filter rules and rule-based classifiers to distinguish between relational and non-relational tables. They filter out small tables, attribute/value tables, and tables embedded inside HTML forms and calendars, eliminating 89.4\% of all tables. In order to train a classifier, they create seven features inspired by Wang and Hu \cite{wang_machine_2002}. While they do not propose novel approaches for web table detection, they are the first to apply the algorithms to a corpus at web scale, containing 14.1 billion HTML tables extracted from the google.com Web crawl. As they optimize for recall of relational tables, they report a F-measure of 73.1\% on a labeled subsample of the data.

Crestan and Pantel \cite{crestan_web-scale_2011} present a supervised framework to classify HTML tables into a more fine-grained table type taxonomy, consisting of nine types, constituting a more challenging classification task. Employing filter rules as a minimum of two rows and columns and no cell containing more than 100 characters about 80\% of sampled tables were filtered out. A set of layout, genuine, and lexical features is proposed and used to train Gradient Boosted Tree classifier. On a dataset containing 5,000 randomly sampled tables, they report an F-measures of 68.3\% for genuine and 89.3\% for layout tables. 

Son and Park \cite{son_web_2013} propose feature generation using the HTML document structure, incorporating the structure within a table as well as the structure appearing in the context of the table. In combination with the features proposed by Wang and Hu, they achieve an F-measure of 98.58\% on their dataset \cite{wang_machine_2002}. 

Lehmberg et al. \cite{lehmberg_mannheim_2015} conduct web table classification in order to create a large open corpus of relational HTML tables, i.e., the WebDataCommons HTML Tables Dataset\footnote{\url{http://webdatacommons.org/webtables/}} based on the 2012 version of the Common Crawl web corpus. They use heuristic filter rules eliminating tables containing nested tables as well as small tables, and employ a classifier using 16 layout, genuine type and word group features, similar to Wang and Hu \cite{wang_machine_2002}. On a manually labeled Gold Standard data set of 7,350 randomly sampled web sites, consisting of 77,630 tables, they report a precision of 58\% and recall of 62\%. Based on this, they created a web table corpus containing 35.7 million tables, which was later updated incorporating contributions by Eberius et al. \cite{eberius_building_2015}. 

Eberius et al. \cite{eberius_building_2015} address the distinction of layout and genuine tables, as well as a more fine-grained classification of the latter. They apply pre-selection Filters, eliminating small tables as well as tables that were invalid or could not be displayed correctly. They incorporated and extended features proposed by Wang and Hu \cite{wang_machine_2002} and Crestan and Pantel \cite{crestan_web-scale_2011}, differentiating between global features, considering the whole table, and local features only computed per row or column. The 29 most relevant features of the initial 127 created features were selected using correlation-based feature selection.
The evaluation of different classifiers is performed on a manually labeled data set containing 1,022 tables. For the best performing classifier, the Random Forest a F-measure value of 95.2\% is reported for the binary classification task. The proposed approach was used to create the Dresden Web Table Corpus \footnote{\url{https://wwwdb.inf.tu-dresden.de/misc/dwtc/}} consisting of millions of web tables.

A similar method can be seen in all of those approaches: Typically, a two-step approach is used, which applies a rough pre-filter for non-genuine tables (usually based on the number of rows and columns and/or nested tables) as a first step. In a second step, a classifier is used to distinguish the remaining tables as genuine or non-genuine.
This is done with the help of combinations of different features which can be divided into different groups \cite{lehmberg_web_2019}. Global features describe the table as a whole, while Local features that are similar to average cell length, which are created for rows or columns individually \cite{crestan_web-scale_2011,eberius_building_2015}.  Another distinction is made between content type features and structural or layout features. The former describe for instance the frequency of certain types of content in cells as images, hyperlinks, forms, alphabetic or numeric characters or empty cells. The latter is for instance represented by the average, variance, minimum or maximum number of rows, columns or cell length~\cite{cafarella_uncovering_2008,crestan_web-scale_2011,eberius_building_2015,wang_machine_2002}.

\subsection{Visual Representation Based Features}
What unites all previous approaches is the approach to generate the features used for classification directly from the HTML source code of the tables. In contrast, only few approaches have been made to directly employ visual properties of web tables for table detection. 

First, Cohen et al. \cite{cohen_flexible_2002} brought up the idea to include visual characteristics by ``quasi-rendering'' HTML source code and using the results to detect relational knowledge on the web. They presented a wrapper learning system, which employs multiple document representations. For the entailed table-based extraction, they define the table detection problem as a binary classification task for the $<$table$>$ tag elements. Two classes of features are extracted, first those originating from the HTML representation of the table and second model based features that are extracted from an abstract rendering of the table in order to represent the two dimensional geometric structure of tables. This abstract rendering is achieved by a geometric table model inferred from processes considering how a HTML may be presented in the browser. On a labeled test sample of 339 tables, an F-measure of 95.9\% is reported.

Another approach to identify and extract information from tables on the web, relying on visual features is taken by Gatterbauer et al. \cite{gatterbauer_towards_2007}.  With the goal of domain independent information extraction from web tables they propose performing table extraction and interpretation on a variant of the CSS2 visual box model as rendered by a web browser, rather than on the source code tree structure (however, the actual style sheet information is not used). In contrast to all other literature considered, Gatterbauer et al. \cite{gatterbauer_towards_2007} do not use HTML $<$table$>$ tag to identify the location of tables, arguing that tables as defined by representing relations reflected by certain visual properties and horizontal and vertical alignment of data in a grid structure become visible after a web page is rendered and are not necessarily defined by the corresponding HTML tag. For the task of table extraction on a data set of 493 web tables a recall 81\% of and a precision of 68\% are reported.

In a recent work, Kim and Hwang proposed a method for table detection in Web pages, which also exploits visual features derived from the image representation of the Web page~\cite{kim2020rule}. The authors report an F1 score of 0.71 and 0.78 on two different test cases.

Although those approaches consider visual features of web tables for the detection of genuine tables, they both rely on rather coarse visual representations and do not use the actual image of the website. This implies that important signals that could be used for the detection (e.g., styling information such as lines or background colors and images, which are not defined in the DOM tree, but in an exterior CSS file) are, so far, not used by any approach. Hence, the approach pursued in this paper is the first approach to exploit the \emph{full} visual appearance of a table, as it works on the rendered image as it would be presented to a human visitor of a web page.


\section{Approach}
\label{sec:approach}
\begin{figure}[t]
    \centering
    \includegraphics[width=\textwidth]{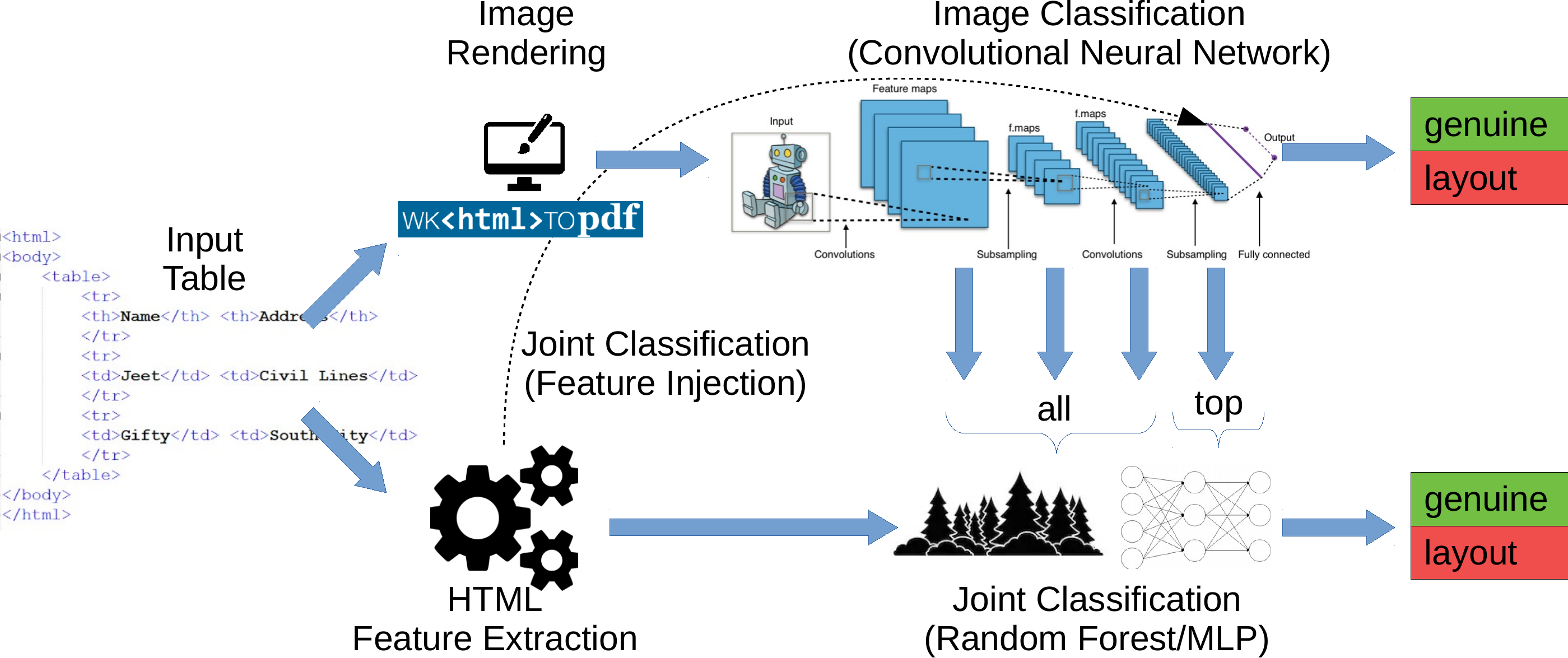}
    \caption{Approach visualized. The upper part shows the purely visual approach. The dashed line and the lower part illustrate the joint approaches.}
    \label{fig:approach}
\end{figure}

The key idea of our approach is that identifying genuine tables on web pages is rather straight forward for human beings, due to the fact that they have certain visual properties that characterize them and distinguish them from other website elements. Hence, it should be possible to also train a machine learning classifier to directly distinguish genuine and layout tables from the image of the rendered HTML table code. However, to obtain table images, only the part defined as a table by the $<$table$>$ tags is rendered. When doing so, the already mentioned style information like CSS and style tags can be used to obtain the exact representation of the table on the website. Additionally, more complex tables structures, as nested tables, which are often filtered out during the filtering step, can be included in the analysis. The only filter rule applied before rendering the pictures is to eliminate obviously non-genuine tables with less than two rows or columns.

In the purely visual approach, the binary classification task is solved using solely a CNN classifier, including a fully connected classification layer. Furthermore, we propose two joint strategies for combining visual and HTML features: (1) injecting HTML as additional features in the fully connected classification layer of the CNN classifier, (2) and combining the features extracted by the CNN and the HTML features in a downstream classifier (here, we use a Random Forest and a MLP classifier). For the latter case, we experiment with extracting only the highest layer features, as well as features from all layers of the CNNs.

Additionally, CNNs are used to extract visual features from the web table images which are then used in a joint approach, using the original DWTC Random Forest classifier \cite{eberius_building_2015} combining HTML features and visual features.

Fig.~\ref{fig:approach} shows the overall approach. The images are rendered in color employing the open source wkhtmltoimage package\footnote{\url{https://github.com/wkhtmltopdf/wkhtmltopdf}}. The rendered images then fed into a convolutional neural network for image classification, leading to a binary classification (\emph{genuine} or \emph{layout}).

\subsection{Convolutional Neural Nets for Image Classification and Visual Feature Extraction}
One of the most commonly used learning algorithms for image classification are Convolutional Neural Networks (CNNs). CNNs are hierarchical feed forward networks, implementing several convolution stages, which are often a combination of convolution layers, non-linear transformation and pooling layers \cite{jarrett_what_2009,lecun_convolutional_2010}. Convolution operations allow the network to  extract relevant features from local correlated data points \cite{khan_survey_2020}. By the stacking of multiple of these feature extraction stages, more abstract representations of the image data are learned subsequently \cite{zheng_good_2016}. When used for classification, the convolution and pooling layers are connected to a classification module consisting of one or several dense layers, emitting classification predictions. Here, the convolutional and pooling layers can bee seen as automatic \emph{feature extractors}, which create the features used by the downstream classifier.

Another way to leverage CNNs' ability to automatically learn representations of visual characteristics in images is to use it as a standalone feature extractor. After the weights have been trained on a certain image classification task, the fully connected layers of the network are removed. Thus, instead of a class prediction, the CNN outputs feature vectors that can be used in other downstream tasks. The use of such extracted visual feature vectors allows the combination with HTML-based features.

Since CNNs subsequently create higher level features, we inspected two different approaches: using only the most abstract representation (the approach depicted in Fig.~\ref{fig:approach} as \emph{top}) and using features from all levels of abstractions (the approach depicted in Fig.~\ref{fig:approach} as \emph{all}).

\subsection{Transfer Learning with VGG16 and ResNet50}
While in classic machine learning, a model for a given classification task is learned from scratch, using labeled examples for that task. In contrast, \emph{transfer learning} refers to re-using a machine learning model trained for a different task, and adapting it for the task at hand \cite{pan2009survey}. In our case, we use re-use generic models for image classification and apply them to the task of table classification.

In our approach, we reuse pre-trained CNNs as a starting point. For adapting (i.e., \emph{transferring}) them to the classification task at hand, we use new input data from our task. This can be accomplished either by freezing the pre-trained weights in the convolution part of the network and allowing only the dense classification layer to adjust weights (we refer to that approach as \emph{frozen}), or by allowing all pre-trained weights in the network to adjust to the new task during training (we refer to that approach as \emph{adapt}). Transfer learning increases efficiency compared to training the weights from scratch and has shown to produce good results. Some well established CNN architectures for the task of image classification are VGG16 and ResNet50  \cite{khan_survey_2020}.

\emph{VGG16} is a deep convolutional neural network, built by stacking several convolutional layers while using small filters of size 3x3 \cite{simonyan_very_2015}. The architecture consists of five convolutional blocks consisting of two to three convolution layers employing ReLu activation functions, followed by a max pooling layer. The output of the convolutions is flattened and send through two dense layers, followed by a classification layer on top. In total it entails 16 trainable layers, of which are 13 convolution layers and three dense layers. In this case of a binary classification task, a classification layer is replaced by a dense layer with one output node and sigmoid activation function is used. When used as feature extractor, global average pooling of the output of convolutional blocks is applied to obtain the feature vector.

\emph{ResNet50}, short for \emph{Residual Network}, is characterized by implementing a residual learning framework, where layers learn residual functions with regard to input layers, in order to facilitate training of very deep networks \cite{he_deep_2015}. Besides implementing identity based skip connections to enable cross layer connectivity, it also incorporates batch normalization as regulating units. The model consists of a total of 50 layers. A first convolution block is followed by four stages of stacked convolution and identity blocks. Similar to the approach for VGG16, we refined the classification layer according to the binary classification task. When the model is used for feature extraction the output after applying global average pooling is used.

For both VGG16 and ResNet50, we use pre-trained models which were built on the ImageNet challenge dataset, an image classification training dataset of more than 1 million images and 1,000 classes. 

\section{Experiments}
\label{sec:experiments}
While most approaches discussed in section~\ref{sec:related} use datasets which are not publicly available, the only previous work using an open dataset is the work using the Dresden Web Table Corpus (DWTC) \cite{eberius_building_2015}. However, that corpus does not contain the original CSS files and images from the Web pages; hence, a rendering of the Web page as it would look in a browser was not possible based on the dataset. Therefore, we manually annotated a new ground truth dataset.

\subsection{Dataset}

\begin{table}[t]
    \footnotesize
    \caption[Gold Standard dataset]{Gold Standard dataset}
    \label{tab:gs}
	\begin{center}
		\begin{tabular}{llll}
				\hline
			& Layout        & Genuine       & Total          \\ \hline
			Training          & 3938          & 4454          & 8392           \\
			Validation     & 1018          & 1080          & 2098           \\
			Test           & 1267          & 1355          & 2622           \\ \hline
			\textbf{Total} & \textbf{6223} & \textbf{6889} & \textbf{13112}
		\end{tabular}
	\end{center}
\end{table}

To collect the data used in this paper, the HTML code of roughly 1.25 million websites was sampled, by accessing 125 randomly chosen WARC files of the March 2020 common crawl\footnote{ \url{https://commoncrawl.org/2020/04/}}. From each of these websites one table, identified by the HTML $<$table$>$ tag, was randomly selected if it contained one. Subsequently, a simple filter was applied, similar to Eberius et al. \cite{eberius_building_2015}. This filter eliminated all tables that had less than two columns or two rows. 
This simple filter rule alone led to an exclusion of about 65\% the sampled tables. 
In addition, the language was restricted to English,using the langdetect\footnote{\url{https://github.com/shuyo/language-detection}} library in Python 
to ensure the interpretability of images during manual ground truthing.

Afterwards, the collected web tables present in the HTML were rendered to image files. To that end, linked images and CSS files were retrieved based on their URL (since they are often not contained in the Common Crawl). To handle cases where the style files were no longer accessible, we tried using the Wayback machine\footnote{\url{https://archive.org/web/}} to download historic snapshots instead, however, this lead to no improvements. Furthermore, we discarded tables that could not be rendered (e.g., due to invalid HTML code). However, the share of discarded tables in that step is only about .01\%.

To label the tables consistently, a precise definition of layout and genuine tables must be established. The tables labeled as genuine tables largely correspond to those defined as relational tables by Crestan and Pantel \cite{crestan_fine-grained_2010}. These include vertical and horizontal lists, matrix and attribute/value tables and enumerations. Two examplary genuine tables can be seen in figures \ref{fig:example_genuine1} and \ref{fig:example_genuine2}. Only forms such as log in elements or address forms were not defined as genuine tables in this data generation. Although they are to be interpreted as relational tables according to Crestan and Pantel \cite{crestan_fine-grained_2010}, they do not contain knowledge to be extracted but, as is the inherent characteristic of a form, blanks. Consequently, they do not serve the purpose of information extraction. Tables classified as layout tables are, for example, formatting tables used to visually arrange contents, as in Fig.~\ref{fig:example_layout1}, or navigational tables used to navigate the website as in Fig.~\ref{fig:example_layout2}.

The resulting gold standard dataset used to develop the classification models presented in this paper contains 13,112 tables, of which 6,889 were labeled as genuine and 6,223 as layout tables, as shown in table \ref{tab:gs}. For these, the original HTML code as well as an image version is present. For model development the data is split into a training set of 80\% and a test set of 20\%, containing 2,622 tables. A 20\% validation set was again split off from the resulting training set, resulting in a validation set of 2,098 tables and a training set of 8,392 tables.\footnote{The code and data used in this paper are available at \url{https://github.com/babettebue/web-table-classification}.}

\subsection{Experimental Setup}
In order to evaluate our approach, we compare different settings to the state of the art classifier used to build the Dresden Web Table Corpus.
\subsubsection{Baselines}
As a baseline for this work, we use the approach presented by Eberius et al. \cite{eberius_building_2015} using a combination of content and layout features extracted from tables' HTML code. The best performing model was implemented in the extraction of the Dresden Web Table Corpus (DWTC), referenced earlier. The DWTC-Extractor\footnote{\url{https://github.com/JulianEberius/dwtc-extractor/}}, containing the full code used for the extraction is available on GitHub. 
Single elements of this extractor, namely the creation of HTML features and the trained WEKA Random Forest classifier for the classification of genuine and layout tables, can be accessed to apply the approach to the newly generated Gold Standard test data. On their own data, Eberius et al. \cite{eberius_building_2015} reported a weighted F1 score of 9.52\%. In addition to applying the model trained on the original dataset, we also retrained the Random Forest classifier on our training dataset.

 
The DWTC extractor additionally filters nested tables before generating the features. Therefore, for the nested tables contained in our Gold Standard, a default layout classification was applied. Regarding the labels of these nested tables reveals that the vast majority indeed are layout tables. 

The hyperparameters of the default DWTC classifier are the standard settings of the Weka Random Forest classifier, i.e., growing 10 trees and no other restrictions for the Random Forest. 
For the retrained Random Forest, trained with scikit-learn\footnote{\url{https://scikit-learn.org/stable/}} in Python, randomized grid search was employed for hyper-parameter tuning, using the described validation set. The best resulting hyperparameter setting was growing 1,600 trees, with a maximal depth of 80, a minimum of 4 samples per leaf, and a minimal number of samples per split of 2.

\subsubsection{CNN Approaches}
The CNN based approaches are implemented using Tensorflow\footnote{\url{https://www.tensorflow.org/}} in Python. The built in functional models of VGG16 and ResNet50, with an option to load weights pre-trained on the ImageNet dataset, were used and adapted. In order to adapt these networks, which are designed for classification on the ImageNet dataset containing 10 image classes, to the binary classification problem. The classification layer was replaced by a dense layer with one output node and a sigmoid activation function. 

Before feeding the images into the CNN, they have to be resized and normalized. The latter is realized by implementing a rescaling layer into the network, rescaling all images to 224x224. First, the models were trained only employing the weights pretrained on the ImageNet dataset. This was achieved by setting all layers but the last dense layer used for classification to non-trainable. This setting is referred to as \emph{frozen}.
In a second configuration, the ImageNet weights were used as initialization weights, which could be fine tuned during the training for our binary classification task. This setting is referred to as \emph{adapt}.

The models were trained for 100 epochs, using the Adam optimizer, binary crossentropy loss and the standard learning rate of .001 for the models with frozen ImageNet weights. The models allowed to fine-tune weights were trained with a smaller learning rate of .0001. A callback with a patience of 20, for frozen weight models and 50 for fine-tuned models, in case of steady validation accuracy was employed to prevent over-fitting. 
Training of VGG16 frozen was stopped after 36 epochs, training of VGG16 adapt after 58 epochs. The frozen ResNet50 was trained for 36 epochs and the fine-tuned ResNet allowed to adapt weights for 69 epochs. 

\subsubsection{Joint Approaches}
For the combined architectures, we used both the extracted HTML features from the feature-based classification approach, as well as the latent features extracted by the CNNs. To extract features generated by the convolutional blocks of the networks, VGG16 and ResNet50 architectures are used without the top dense layers. For the VGG16 a GlobalAveragePooling layer is stacked on top of the convolutional network in order to obtain a one dimensional feature vector with reduced dimensionality \cite{zheng_good_2016}. The VGG16 outputs 512 visual features. The ResNet50 model already implements a GlobalAveragePooling layer after the last convolution, which outputs feature vector of length 2,048. For both architectures, models with the ImageNet weights, as well as the models fine-tuned for the binary classification problem at hand, are used for feature extraction.


Additionally, for both models features of all levels of abstractions, referred to as \textit{all}, are extracted, GlobalAveragePooling is applied and followed by a simple concatenation of the feature vectors.  For VGG16 this results in a feature set containing 1,472 visual features and for ResNet50 in a set of 3,903 features.

\subsection{Results}
The recall, precision, and F1 score both on the layout and genuine table class, as well as the weighted average of those, are shown in table~\ref{tab:results}.

We can observe that while the original DWTC classifier does not perform optimally on the dataset at hand, retraining the classifier works a lot better. The results of the retrained classifier slightly outperform the results reported in the original paper, where a weighted average F1 score of .906 was reported~\cite{eberius_building_2015}.

Applying the \emph{frozen} VGG16 and ResNet50 architectures without weight adaption yields results below the baseline. On the other hand, allowing fine-tuning of weights in the \emph{adapt} setting, the results are in a similar range as the retrained DWTC classifier, with ResNet50 even outperforming those results by a small margin. The best joint approaches combining both the DWTC and the CNN features again perform a little better than the purely visual and the DWTC approach alone.

\begin{table}[t]
    \footnotesize
    \caption{Results with baselines, visual classification approaches (upper part), and joint approaches based on feature injection, RandomForest (RF), and MLP classifiers (lower part)}
    \label{tab:results}
    \centering
    \begin{tabular}{l|r|r|r||r|r|r||r|r|r}
    &   \multicolumn{3}{c||}{Layout}&   \multicolumn{3}{c||}{Genuine}&   \multicolumn{3}{c}{Weighted avg.}\\
    Approach & P & R & F1 & P & R & F1 & P & R & F1 \\
    \hline
    DWTC original	&	.894	&	.916	&	.889	&	.917	&	.865	&	.890	&	.891	&	.890	&	.890 \\
    DWTC retrained	&	.934	&	.924	&	.929	&	\textbf{.930}	&	.939	&	.934	&	.932	&	.932	&	.932 \\
    \hline
    VGG16 frozen    &   .901    &   .824    &   .861    &   .848    &   .916    &   .880    &   .873    &   .871 &   .871 \\
    VGG16 adapt    &   .915    &   .925    &   .920    &   .929    &   .920    &   .924    &   .922    &   .922    &   .922\\
    ResNet50 frozen    &   .914    &   .875    &   .894    &   .887    &   .923    &   .905    &   .900    &   .900    &   .900\\
    ResNet50 adapt    &   \textbf{.937}    &   \textbf{.929}    &   \textbf{.930}    &   .929    &   \textbf{.942}    &   \textbf{.936}    &   \textbf{.933}    &   \textbf{.933}    &   \textbf{.933}\\
    \hline
    \hline
    Injection VGG16 frozen &  .892	&	.921	&	.906	&	.924	&	.895	&	.909	&	.908	&	.908	&	.908\\
    Injection VGG16 adapt & .878	&	.913	&	.895	&	.916	&	.881	&	.898	&	.897	&	.897	&	.897\\
    Injection ResNet50 frozen & .821	&	.853	&	.837	&	.857	&	.826	&	.841	&	.840	&	.839	&	.839\\
    Injection ResNet50 adapt & .854	&	.766	&	.807	&	.800	&	.877	&	.837	&	.826	&	.823	&	.823\\
    \hline
    RF VGG16 frozen (top)   &   .945   &   .923   &   .933   &   .930   &   .949   &   .939   &   .936   &   .936   &   .936\\
    RF VGG16 frozen (all)   &   .943   &   .925   &   .934   &   .931   &   .949   &   \textbf{.940}   &   \textbf{.937}   &   \textbf{.937}   &   \textbf{.937}\\
    RF VGG16 adapt (top)   &   .938   &   .913   &   .925   &   .921   &   .943   &   .932   &   .929   &   .929   &   .929\\
    RF VGG16 adapt (all)   &   .931   &   .918   &   .925   &   .924   &   .937   &   .930   &   .928   &   .928   &   .928\\
    RF ResNet50 frozen (top)   &   .938   &   .922   &   .930   &   .928   &   .943   &   .937   &   .933   &   .933   &   .933\\
    RF ResNet50 frozen (all)   &   .937   &   .926   &   \textbf{.937}   & .931   &   .942   &   .937   &   .934   &   .934   &   .934\\
    RF ResNet50 adapt (top)   &   .930   &   .916   &   .923   &   .923   &   .936   &   .929   &   .926   &   .926   &   .926\\
    RF ResNet50 adapt (all)   &   .927   &   .920   &   .923   &   .925   &   .932   &   .929   &   .926   &   .926   &   .926\\
    \hline
    MLP Joint VGG16 frozen (top) & .911	&	.927	&	.919	&	.931	&	.915	&	.923	&	.921	&	.921	&	.921\\
    MLP Joint VGG16 frozen (all) & .922	&	.891	&	.906	&	.901	&	.929	&	.915	&	.911	&	.911	&	.911\\
    MLP Joint VGG16 adapt (top) & .875	&	\textbf{.935}	&	.904	&	\textbf{.935}	&	.875	&	.904	&	.906	&	.904	&	.904\\
    MLP Joint VGG16 adapt (all) & .907	&	.912	&	.909	&	.917	&	.912	&	.915	&	.912	&	.912	&	.912\\
    MLP Joint ResNet50 frozen (top) & .914	&	.900	&	.907	&	.908	&	.921	&	.914	&	.911	&	.911	&	.911\\
    MLP Joint ResNet50 frozen (all) & .910	&	.931	&	.920	&	.934	&	.914	&	.924	&	.922	&	.922	&	.922\\
    MLP Joint ResNet50 adapt (top) & \textbf{.949}	&	.509	&	.663	&	.680	&	\textbf{.974}	&	.801	&	.810	&	.749	&	.734\\
    MLP Joint ResNet50 adapt (all) & .918	&	.595	&	.722	&	.715	&	.951	&	.816	&	.813	&	.779	&	.771\\
\end{tabular}
\end{table}

McNemar's non parametric test \cite{everitt_analysis_1977} was conducted to compare the best performing classifiers from each approach, as recommended by Dietterich et al. \cite{diett_1998} if evaluation is performed on a single test set. 
The results reveal that there is \emph{no} significant difference between the retrained DWTC classifier, the best performing CNN classifier. Hence, our conclusion is that both the visual approach and the approach based on explicit feature engineering work equally well.

\begin{figure}[t]
    \centering
    \begin{subfigure}[b]{.32\textwidth}
        \includegraphics[width=\textwidth]{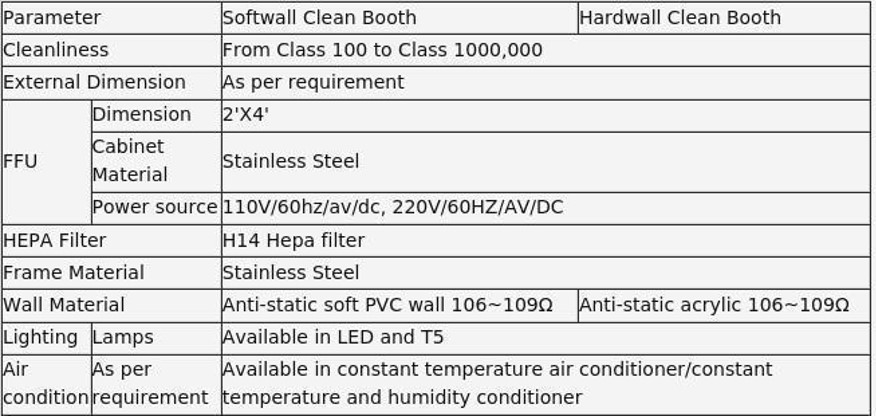}
        \caption{Nested table}
    \end{subfigure}
    \begin{subfigure}[b]{.32\textwidth}
        \includegraphics[width=\textwidth]{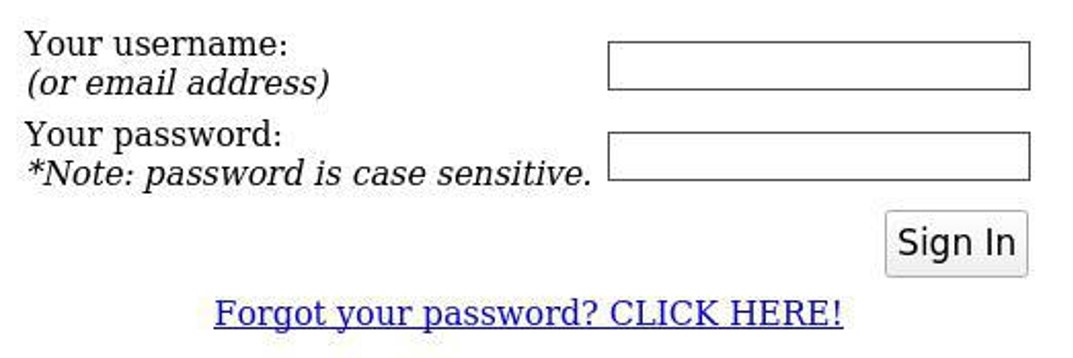}
        \caption{Form}
    \end{subfigure}
    \begin{subfigure}[b]{.32\textwidth}
        \includegraphics[width=\textwidth]{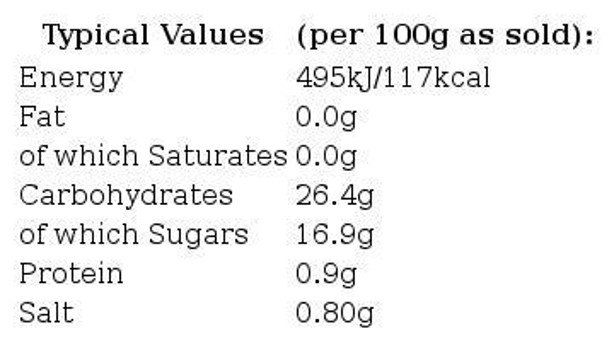}
        \caption{Table without lines}
    \end{subfigure}

    \caption{Examples for misclassified tables}
    \label{fig:examples_mistakes}
\end{figure}

While the results are very different in terms of overall performance, they reveal striking difference when looking at the mistakes they make. Fig.~\ref{fig:examples_mistakes} shows a few typical mistakes. More complex tables like the one shown in (a) are often misclassified by the DWTC approach, but handled correctly by the CNN-based classifiers. On the other hand, (b) and (c) show two typical mistakes made by the CNN-based classifiers: they tend to misclassify tables for layouting input forms as genuine tables, and often do not recognize tables without lines as genuine tables. This shows that vertical and horizontal lines are features which have a very strong importance for the CNN-based classifiers.

Another interesting observation is that when using the CNN architectures alone, the approaches based on adapting weights outperform those without, and ResNet50 outperforms VGG16. On the other hand, in the joint approaches, the trend is reversed: adapted approaches work worse, and the best results are achieved with VGG16. One possible explanation is that when fine-tuning the weights to the task of table classification, the features that are learned by the CNN become more similar to the explicitly created ones, and therefore, the information gain by combining the extracted features with the explicitly created ones is smaller. Moreover, being a smaller model, VGG16 might have a tendency to extract more coarse grained features which are less overlapping with the HTML based features.

\section{Conclusion and Outlook}
\label{sec:conclusion}
In this paper, we have introduced a visual classification approach for distinguishing tables on the Web, in particular, genuine and layout tables. The results show that the purely visual approach yields results which are of the same quality as the current state of the art, which is based on extracting explicit features from the HTML code. We conclude that purely visual approaches are a suitable alternative to the state of the art, since they are also more versatile, as they can handle information defined in style sheets, dynamically built Web pages, etc.

An in-depth inspection of the results has revealed that the mistakes made by the approach based on HTML features and the visual approaches are different. This raises the assumption that joint approaches could yield even better results, however, our results so far did not show a significant improvement. Moreover, using other pre-trained image models more tailored to the task, like \emph{TableNet}~\cite{paliwal2019tablenet}, might improve the results.

So far, we have only considered one task, i.e., the distinction of layout and genuine tables. While the approach could be transferred to other tasks, such as a finer-grained distinction of different table types \cite{crestan_web-scale_2011,eberius_building_2015}, experimental results are still outstanding.

Web table classification is not the only task in Web information extraction where visual signals can be exploited.
In the future, we plan to evaluate whether visual approaches can also be used for detecting certain content elements on a Web page which have a common visual appeal, such as addresses or opening hours. Since such elements are often marked up with Microdata or Microformat annotations, training data for such approaches could easily be sourced \cite{meusel2014webdatacommons}. Here, visual approaches could also help in building information extraction systems which work on HTML data.

Another interesting field is the classification of entire Web pages. Since we assume that news pages, e-commerce pages, discussion pages etc. can also be identified based on certain visual signals, visual approaches could also be of interest here. Current works is usually based on images on the Web page, but not rendered images of HTML content \cite{hashemi2020web}.

\subsection*{Acknowledgements}
We would like to thank Julius Gonsior and Maik Thiele at TU Dresden for their assistance in accessing the DWTC dataset and classifier for the experiments.

\bibliographystyle{splncs04}
\bibliography{bibliography}

\end{document}